\documentclass[a4paper,twoside]{article}
\usepackage{epsfig}
\usepackage{subcaption}
\usepackage{calc}
\usepackage{amssymb}
\usepackage{amstext}
\usepackage{amsmath}
\usepackage{amsthm}
\usepackage{multicol}
\usepackage{pslatex}
\usepackage{apalike}
\usepackage{algorithm2e}
\usepackage[bottom]{footmisc}
\usepackage{tikz}
\usetikzlibrary {shapes.geometric} 
\usepackage{booktabs,makecell,tabularx}
\usepackage{hhline}
\usepackage{multirow}
\usepackage{adjustbox}
\usepackage[binary-units]{siunitx}
\usepackage{pifont}

\usepackage[acronym,toc]{glossaries}

\usepackage{hyperref}

\makeglossaries
\newacronym[firstplural=concept drift detectors]{dd}{DD}{concept drift detector}
\newacronym{fp}{FP}{False Positive}
\newacronym{fn}{FN}{False Negative}
\newacronym{dl}{DL}{deep learning}
\newacronym{ml}{ML}{machine learning}
\newacronym{moa}{MOA}{Massive Online Analysis}
\newacronym{bdcp}{BDCP}{Beta Distribution Change Point}
\newacronym{iks}{IKS}{Incremental Kolmogorov-Smirnov}

\usepackage{SCITEPRESS}
 
\begin{document}

\title{Towards Computational Performance Engineering for \\ Unsupervised Concept Drift Detection - \\ Complexities, Benchmarking, Performance Analysis}

\author{\authorname{Elias Werner\sup{1,2}\orcidAuthor{0000-0003-2549-7626}, Nishant Kumar\sup{2,3}\orcidAuthor{0000-0001-6684-2890}, Matthias Lieber\sup{1,4}\orcidAuthor{0000-0003-3137-0648}, Sunna Torge\sup{1,2}\orcidAuthor{0000-0001-9756-6390}, Stefan Gumhold\sup{1,2,3}\orcidAuthor{0000-0003-2467-5734}, Wolfgang E. Nagel\sup{1,2,4}}
\affiliation{\sup{1}Center for Interdisciplinary Digital Sciences (CIDS), Technische Universität Dresden, Germany}
\affiliation{\sup{2}Center for Scalable Data Analytics and Artificial Intelligence (ScaDS.AI) Dresden/Leipzig. Germany}
\affiliation{\sup{3}Chair of Computer Graphics and Visualization (CGV), Technische Universität Dresden, Germany}
\affiliation{\sup{4}Center for Information Services and High Performance Computing (ZIH), Technische Universität Dresden, Germany}
\email{\{elias.werner, nishant.kumar, matthias.lieber, sunna.torge, stefan.gumhold, wolfgang.nagel\}@tu-dresden.de}
}


\abstract{Concept drift detection is crucial for many AI systems to ensure the system's reliability. These systems often have to deal with large amounts of data or react in real-time. Thus, drift detectors must meet computational requirements or constraints with a comprehensive performance evaluation. However, so far, the focus of developing drift detectors is on inference quality, e.g.~accuracy, but not on computational performance, such as runtime.
Many of the previous works consider computational performance only as a secondary objective and do not have a benchmark for such evaluation.
Hence, we propose and explain performance engineering for unsupervised concept drift detection that reflects on computational complexities, benchmarking, and performance analysis. We provide the computational complexities of existing unsupervised drift detectors and discuss why further computational performance investigations are required. Hence, we state and substantiate the aspects of a benchmark for unsupervised drift detection reflecting on inference quality and computational performance. Furthermore, we demonstrate performance analysis practices that have proven their effectiveness in High-Performance Computing, by tracing two drift detectors and displaying their performance data.
}

\onecolumn 
\maketitle 
\normalsize 
\setcounter{footnote}{0} 
\vfill

\section{\uppercase{Introduction}}
In the last years, the amount of available data increased significantly and is expected to be about \SI{175}{Z\byte} only for the year 2025~\cite{reinsel2018data}. The availability of vast amounts of data and the exploitation of computing resources such as graphics processing units (GPUs) or tensor processing units (TPUs) led to the emergence of \gls{dl} methods in many applications as predictive maintenance~\cite{martinez2018labelling}, marine photography~\cite{langenkamper2020gear}, transportation planning~\cite{grubitzsch2021ai} or computer vision tasks such as object detection~\cite{kumar2023normalizing}.

However, many of these applications face different challenges which concern 1.) the inference quality (e.g. accuracy) of the model and thus the effectiveness of the application and 2.) the computational performance of the application in terms of the compute time and the compute resources.

The effectiveness of the applications is often determined by the inference quality (e.g. accuracy) of the \gls{dl} model on a different data distribution than the distribution which the model was trained with.
However, while pure \gls{dl} based applications work nicely on the training data distribution, they might not perform as well when the test data distribution is different from the training data distribution. Such changes in the data distributions are referred to as concept drift and are documented in many different application fields.
For example, \cite{grubitzsch2021ai} outlined that the robustness of AI models is questionable for sensor data-based transport mode recognition. The reason is the variety of context information, e.g. device type or user behavior that introduces concept drift into the data.
\cite{langenkamper2020gear} demonstrated concept drift when using different gear or changing positions in marine photography and explained the effect on \gls{dl} models.
Hence, such applications need to be accompanied by approaches such as \gls{dd} to estimate changes in the data distribution and to decide the robustness of a \gls{dl} model on a given input. \glspl{dd} that require the immediate availability of data labels are referred to as supervised and \glspl{dd} that reduce the amount of required data labels or can operate completely in the absence of labeled data are referred to as unsupervised.

The second challenge for applications is to handle large amounts of data or high-speed data streams and react in real-time.~On the other hand, applications are often bound to certain hardware requirements or have to operate with limited computational resources. These observations should point to the necessity of thorough investigations concerning computational performance, i.e.~runtime, memory usage, and scalability. Thus, with \glspl{dd} being an important part of a robust \gls{dl} pipeline, it is also necessary to conduct such investigations on \glspl{dd}. Note that we refer to metrics as runtime or memory usage as \textit{computational performance} and to metrics such as accuracy or recall as \textit{inference quality}. 
However, the literature does not focus on computational performance investigations but concentrates on the methodological improvements and inference quality evaluation of \glspl{dd} on small-scale examples only as outlined in the survey by~\cite{gemaque2020overview}. 
Moreover, while theoretical computational complexities play a crucial role in understanding algorithmic behaviors, they fall short of encompassing the real-world performance of an algorithm. This is attributed to various factors such as implementation, compiler optimizations, data distribution, and other external influences, which become significant when the algorithm operates on real hardware and processes real data.

Our work focuses on unsupervised \glspl{dd} that can operate in the absence of data labels, as the provision of labeled training data is very expensive and for many applications not given. 
~So far, there is no previous work comprehensively investigating the computational performance of unsupervised \glspl{dd}. In response, our work reflects on the field of performance engineering, which has its roots in High-Performance Computing (HPC) and introduces it for unsupervised \glspl{dd}. Thus, we discuss important pillars for a comprehensive consideration of computational performance: complexity analysis, benchmarking, and performance analysis. 
This work's key contributions include:
\begin{enumerate}
    \item We reflect on the existing literature and show that it lacks computational performance evaluation for unsupervised concept drift detection.
    \item We show a concrete path towards performance engineering of \glspl{dd} and discuss complexity analysis, benchmarking, and performance analysis.
    \item We provide the time and space complexities for a set of existing \glspl{dd}.
    \item We state and substantiate the requirements for a benchmark of unsupervised \gls{dd}.
    \item We demonstrate the effectiveness of performance analysis tools by tracing two \glspl{dd} and present the performance data.
\end{enumerate}

The rest of the paper consists of four parts. In Section 2, we introduce preliminaries and define concept drift.~\autoref{sec:unsupervised}, provides an overview of the prior works for unsupervised concept drift detection and explains the scope of previous computational performance evaluation.~In \autoref{sec:compCons}, we introduce performance engineering for unsupervised concept drift detection. We provide computational complexities and discuss them. Furthermore, we examine aspects for a comprehensive benchmark and give initial insights into performance analysis for \glspl{dd}.

\section{\uppercase{Background}}
\label{sec:background}
This section formally defines concept drift and introduces supervised and unsupervised drift detection.

We follow the formal notations and definitions by~\cite{webb2016characterizing} for the following illustrative equations, but also consider~\cite{gama2014survey} and \cite{hoens2012learning} among others. Note that our assumptions hold for the discrete and continuous realms in principle. Nevertheless, for ease of simplification, we consider only the discrete realm in our notations.
Assume for a \gls{ml} problem there is a random variable $X$ over vectors of input features $[X_0, X_1,...,X_n]$. Moreover, there is a random variable $Y$ over the output that can be either discrete (for classification tasks) or continuous (for regression tasks). In this case, $P(X)$ and $P(Y)$ represent the probability distribution over $X$ and $Y$ respectively (priori). $P(X,Y)$ represents the joint probability distribution over $X$ and $Y$ and refers to a concept. At a particular time $t$, a concept can now be denoted as $P_t(X,Y)$.
Concept drift refers to the change of an underlying probability distribution of a random variable over time. Formally:
\begin{equation}
\label{eq:conceptDrift}
P_t(X,Y) \neq {P}_{t+1}(X,Y)
\end{equation}

\noindent Supervised drift detection is the process where the data labels $Y$ are always immediately available and unsupervised \glspl{dd} detect drift without labeled data. 

\section{\uppercase{Related Work}}
\label{sec:unsupervised}
\label{sec:unsupervisedDD}

In the field of supervised drift detection, different investigations by \cite{barros2018large} and \cite{palli2022experimental} exist, that present and compare multiple supervised \glspl{dd}.~Additionally, \cite{mahgoub2022benchmarking} presents a benchmark of supervised \glspl{dd} that considers the runtime and memory usage of the related \glspl{dd} besides the \glspl{dd}' quality.

To the best of our knowledge, there is no such benchmark for unsupervised \glspl{dd}, Nevertheless, surveys exist that summarize the related work. \cite{gemaque2020overview} and \cite{shen2023unsupervised} provide overviews and taxonomies to classify unsupervised drift detectors following different criteria. Although both surveys mention the importance of computational performance considerations, they did not incorporate such objectives in their overviews thoroughly.

Investigating prior methods for unsupervised \glspl{dd} does also not provide enough evidence for thorough computational performance investigations. Several works~\cite{mustafa2017unsupervised,kifer2004detecting,ditzler2011hellinger,haque2016efficient,sethi2017reliable,kim2017efficient,zheng2019labellessShort,cerqueira2022studd,lughofer2016recognizing,deMello2019learning,gozuaccik2019unsupervised,gozuaccik2021concept} do not conduct any runtime, memory, energy or scalability performance measurements.~Thus, it is difficult to assess their computational performance in real-world applications. 
Other works by~\cite{dasu2006information,gu2016concept,lu2014concept,qahtan2015pca,liu2017regional,liu2018accumulating,song2007statistical,dos2016fast,greco2021drift,pinage2020drift} conduct few experiments concerning runtime on multiple datasets and compared their approaches to other works sporadically.~Only \cite{liu2018accumulating} investigated memory utilization. 
~In addition, not all papers perform their evaluation with the same setting and vary in the data sets chosen, the number and dimension of data points, and how the computational performance measurements are carried out.
Although some works consider the theoretical computational complexity of their algorithms, this can not replace empirical measurements on real datasets and machines, e.g. as shown by~\cite{jin2012understanding} highlighting the impact of computational performance bugs on the runtime of implementations.~However, as discussed by \cite{lukats2023reproducible}, source code is often not available and the reproducibility of presented experiments is often questionable.

Future applications with high data volumes, high data velocities, or computational resource constraints will require resource-efficient approaches and implementations.~Contrarily, computational performance aspects for unsupervised concept drift detection were only investigated as a secondary objective in the literature.
Thus, we require proper and well-documented implementations that enable consistent computational performance evaluations to assess the applicability of \glspl{dd} for use cases with high volumes and high-velocity data. 
~Moreover, to obtain resource-efficient AI systems and to avoid waste of resources, scalable or parallel \glspl{dd} and the resource-efficient deployment of the approaches need to be investigated.
For such developments, the HPC community has broad expertise in the field of performance engineering. Besides computational complexity analysis and substantiated best practices in benchmarking, several tools such as Score-P~\cite{knupfer2012score} or Vampir~\cite{vampir} are developed to support the systematic performance analysis of applications. Applying these methods to unsupervised \glspl{dd} supports the evaluation of the computational performance of different approaches and allows systematically setting up resource-efficient solutions for future applications.

\section{\uppercase{Computational~Performance~Engineering}}
\label{sec:compCons}
In this section, we introduce our contributions to assess the computational performance of unsupervised \glspl{dd}. This includes 1.)~the computational complexity of a set of \glspl{dd} in Section 4.1, 2.)~the concept of a comprehensive benchmark in Section 4.2, and 3.)~a clear workflow for performance analysis of \glspl{dd} in Section 4.3, exemplarily presented on two instrumented implementations.


As a motivating example, we randomly selected the two unsupervised \glspl{dd} Incremental Kolmogorov-Smirnov (IKS)~\cite{dos2016fast} and the student-teacher approach STUDD~\cite{cerqueira2022studd} and measured the runtime and memory along with the accuracy and amount of requested labels of four pipelines: 
1.)~IKS:~retraining of the base \gls{ml} model after drift detection with IKS
2.)~STUDD:~retraining of the base \gls{ml} model after drift detection with STUDD
3.)~Baseline 1 (BL1):~pipeline without re-training of the base \gls{ml} model
4.)~Baseline 2 (BL2):~retraining of the base \gls{ml} model after every 5000 samples.
~IKS detects drift based on the changes in the raw input data distribution by continuously applying a Kolmogorov-Smirnov test on a reference dataset and a detection dataset. STUDD consists of a student auxiliary \gls{ml} model to mimic the behavior of a primary teacher decision \gls{ml} model. Drift is detected if the mimicking loss of the student model changes with respect to the teacher's predictions.
All pipelines are implemented in Python and use a Random Forest with 100 decision trees as the base classifier. The base classifier is trained with the first 5000 samples of the dataset, the remaining data is processed as a stream by the several pipelines.
We applied all pipelines on the Forest Covertype~\cite{blackard1999comparative} dataset that consists of 54 features with seven different forest cover type designations in $5.8*10^6$ data samples.
All experiments ran with a single CPU core of an AMD EPYC 7702, fixed to 2.0 GHz frequency.
Note that since the implementations do not run in parallel, we do not consider multiple CPU cores. However, using HPC still offers advantages for benchmarking over a local setup, such as a fixed CPU frequency and low system noise. In addition, we refer to best practices and methodologies in the field of HPC in this section.
~The source code of the experiments for the depicted results will be available publicly~\footnote{\href{https://github.com/elwer/Perf\_DD}{https://github.com/elwer/Perf\_DD}}.

\begin{figure}
    \begin{tikzpicture}[yscale=0.8]
    \draw (0,0) rectangle (0.5,3.4);
    \node[above, anchor=south east] at (0.5,3.4) {\tiny{85}};
    
    \draw (0.5,0) rectangle (1,2.72);
    \node[above, anchor=south east] at (1,2.72) {\tiny{68}};
    
    \draw (1,0) rectangle (1.5,2.44);
    \node[above, anchor=south east] at (1.5,2.44) {\tiny{61}};
    
    \draw (1.5,0) rectangle (2,3.2);
    \node[above, anchor=south east] at (2,3.2) {\tiny{80}};
    
    \draw[->] (-0.5,0) -- (2.5,0) node[right] {};
    \draw[->] (-0.01,-0.5) -- (-0.01,4) node[above] {$100$};
    
    \node[below] at (0.25, 0) {\tiny{\rotatebox{-45}{$IKS$}}};
    \node[below] at (0.75, 0) {\tiny{\rotatebox{-45}{$STUDD$}}};
    \node[below] at (1.25, 0) {\tiny{\rotatebox{-45}{$BL1$}}};
    \node[below] at (1.75, 0) {\tiny{\rotatebox{-45}{$BL2$}}};
    
    \node[above, text width=2cm] at (1.5, 4) {Accuracy ($\%$)};

    \foreach \y in {25,50,75}
        \pgfmathsetmacro{\value}{\y/100*4}
        \draw (0.1,\value) -- (-0.1,\value) node[left] {\y};
\end{tikzpicture}
\hfill
\begin{tikzpicture}[yscale=0.8]
    \draw (0,0) rectangle (0.5,2);
    \node[above, anchor=south east] at (0.5,2) {\tiny{50}};
    
    \draw (0.5,0) rectangle (1,0.12);
    \node[above, anchor=south east] at (1,0.12) {\tiny{3}};
    
    \draw (1,0) rectangle (1.5,0);
    \node[above, anchor=south east] at (1.5,0) {\tiny{0}};
    
    \draw (1.5,0) rectangle (2,4);
    \node[above, anchor=south east] at (2,3.6) {\tiny{100}};
    
    \draw[->] (-0.5,0) -- (2.5,0) node[right] {};
    \draw[->] (-0.01,-0.5) -- (-0.01,4) node[above] {$100$};
    
    \node[below] at (0.25, 0) {\tiny{\rotatebox{-45}{$IKS$}}};
    \node[below] at (0.75, 0) {\tiny{\rotatebox{-45}{$STUDD$}}};
    \node[below] at (1.25, 0) {\tiny{\rotatebox{-45}{$BL1$}}};
    \node[below] at (1.75, 0) {\tiny{\rotatebox{-45}{$BL2$}}};

    \node[above, text width=1.6cm] at (1.5, 4) {Labels ($\%$)};

    \foreach \y in {25,50,75}
        \pgfmathsetmacro{\value}{\y/100*4}
        \draw (0.1,\value) -- (-0.1,\value) node[left] {\y};
    
\end{tikzpicture}
\begin{tikzpicture}[yscale=0.8]
    \draw (0,0) rectangle (0.5,1.75);
    \node[above, anchor=south east] at (0.6,1.7) {\tiny{5266}};
    
    \draw (0.5,0) rectangle (1,3.4);
    \node[above, anchor=south east] at (1.1,3.4) {\tiny{10218}};
    
    \draw (1,0) rectangle (1.5,1.44);
    \node[above, anchor=south east] at (1.6,1.44) {\tiny{4321}};
    
    \draw (1.5,0) rectangle (2,1.48);
    \node[above, anchor=south east] at (2.1,1.48) {\tiny{4432}};
    
    \draw[->] (-0.5,0) -- (2.5,0) node[right] {};
    \draw[->] (-0.01,-0.5) -- (-0.01,4) node[above] {$12k$};
    
    \node[below] at (0.25, 0) {\tiny{\rotatebox{-45}{$IKS$}}};
    \node[below] at (0.75, 0) {\tiny{\rotatebox{-45}{$STUDD$}}};
    \node[below] at (1.25, 0) {\tiny{\rotatebox{-45}{$BL1$}}};
    \node[below] at (1.75, 0) {\tiny{\rotatebox{-45}{$BL2$}}};
    
    \node[above, text width=2.5cm] at (1.7, 3.8) {Runtime(s)};

    \foreach \y in {3,6,9}
        \pgfmathsetmacro{\value}{\y*1000/12000*4}
        \draw (0.1,\value) -- (-0.1,\value) node[left] {\y k};
    
\end{tikzpicture}
\hfill
\begin{tikzpicture}[yscale=0.8]
    \draw (0,0) rectangle (0.5,2.35);
    \node[above, anchor=south east] at (0.5,2.35) {\tiny{704}};
    
    \draw (0.5,0) rectangle (1,3.36);
    \node[above, anchor=south east] at (1,3.36) {\tiny{1008}};
    
    \draw (1,0) rectangle (1.5,1.48);
    \node[above, anchor=south east] at (1.5,1.48) {\tiny{446}};
    
    \draw (1.5,0) rectangle (2,2.33);
    \node[above, anchor=south east] at (2,2.33) {\tiny{698}};
    
    \draw[->] (-0.5,0) -- (2.5,0) node[right] {};
    \draw[->] (-0.01,-0.5) -- (-0.01,4) node[above] {$1.2k$};
    
    \node[below] at (0.25, 0) {\tiny{\rotatebox{-45}{$IKS$}}};
    \node[below] at (0.75, 0) {\tiny{\rotatebox{-45}{$STUDD$}}};
    \node[below] at (1.25, 0) {\tiny{\rotatebox{-45}{$BL1$}}};
    \node[below] at (1.75, 0) {\tiny{\rotatebox{-45}{$BL2$}}};
    
    \node[above, text width=1.6cm] at (1.5, 3.8) {Memory(MB)};

    \foreach \y in {300,600,900}
        \pgfmathsetmacro{\value}{\y/1200*4}
        \draw (0.1,\value) -- (-0.1,\value) node[left] {\y};
    
\end{tikzpicture}
    \caption{Accuracy, amount of labels, runtime, and peak memory of the pipelines IKS, STUDD, BL1, and BL2 on Forest Covertype dataset.}
    \label{fig:example}
\end{figure}

Our results are depicted in~\autoref{fig:example}.~BL1 has the lowest computational demand on the dataset, i.e.~lowest runtime and peak memory. However, it achieves the lowest accuracy without requiring any further data labels while processing the stream.
~BL2 has a slightly higher computational demand than BL1 requests all data labels due to the continuous re-training of the base model and achieves much higher accuracy than BL1.
IKS consumes slightly more memory and runtime than BL2.~It achieves slightly higher accuracy than BL2 while reducing the amount of required labels by 50$\%$.
STUDD requires more memory than the other pipelines and also the runtime is by far the highest.~It achieves a higher accuracy than BL1 but lower than BL2.~However, the strength of STUDD is that the amount of requested labels is reduced to only 3$\%$ of the total stream.
~Hence, we conclude that for our setting, both \glspl{dd} have their strengths: IKS leads to a moderate reduction of requested labels by preserving high accuracy without high computational overhead.~STUDD dramatically reduces the number of requested labels by improving accuracy over BL1 at a high computational cost.

However, based on those measurements, it is unclear how \gls{dd} based pipelines behave in different settings since they highly rely on the chosen datasets, hyperparameters, or hardware constraints.~\cite{barros2018large} demonstrated different behaviors of supervised \glspl{dd} on different datasets.
Thus, we need also a benchmark comprising different unsupervised \glspl{dd}. 
The Section 4.1. aims at an assessment of the scaling behavior in space and time, and the complexity results support the evaluation of the computational performance of \glspl{dd} in a benchmark.

\subsection{Complexity Analysis}


We determined the complexities for a set of \glspl{dd} that do not rely on a \gls{ml} model or another method, that is highly dependent on the underlying use case or data. We determined the complexities based on the mathematical foundations and pseudo-code presented in the original works. Additionally, we aligned complexities that have been presented in the literature according to our notations. Finally, we state the complexities for the approaches \textit{KL~\cite{dasu2006information}, PR~\cite{gu2016concept}, CD~\cite{qahtan2015pca}, NM-DDM~\cite{mustafa2017unsupervised}, HDDDM~\cite{ditzler2011hellinger}, IKS~\cite{dos2016fast}, BNDM~\cite{xuan2020bayesian}, ECHO~\cite{haque2016efficient}, OMV-PHT~\cite{lughofer2016recognizing}, DbDDA~\cite{kim2017efficient}, CM~\cite{lu2014concept}}.~See the Appendix for our justifications of the outlined complexities. Although most approaches require an initialization phase for their drift detection, the amount of data computed in this phase is much smaller than the amount of data processing the stream.~Therefore, we skip the complexities of the initialization phase and focus on the complexity per sample while processing the stream.

\setlength{\tabcolsep}{4pt}
\begin{table}[t!]
\caption{
~Time and Space complexity per sample over the data stream. Parameters: \textit{d}: dimensions, \textit{w}: window size, i.e examined data points, \textit{b}: bins in histogram (NM-DDM and OMV-PHT only), \textit{e}: size of ensemble (DbDDA only), \textit{s}: sampling times (NN-DVI only), \textit{$\delta$}: minimum value for the box length property (KL only).
\vspace{0.1cm}\linebreak
\textbf{Approaches:} KL~\cite{dasu2006information}, PR~\cite{gu2016concept}, CD~\cite{qahtan2015pca}, \textit{NM-DDM}~\cite{mustafa2017unsupervised}, HDDDM~\cite{ditzler2011hellinger}, IKS~\cite{dos2016fast}, BNDM~\cite{xuan2020bayesian}, ECHO~\cite{haque2016efficient}, OMV-PHT~\cite{lughofer2016recognizing}, \textit{DbDDA}~\cite{kim2017efficient}, CM~\cite{lu2014concept}
}

  \renewcommand{\arraystretch}{1.3}
\begin{tabular}{p{1.6cm} p{2.5cm} p{2.6cm}}
\hline
 Approach  & Time Complexity & Space Complexity \\\hline\hline
 KL & $O(d \times \log{(\frac{1}{\sigma})})$ & $O(w \times d)$ \\\hline  
 PR & $O(\log{}^2 w)$ & $O(w \times d)$ \\\hline  
 CD & \makecell[lt]{$O(d)$ \\ \textit{\scriptsize drift:} $O(d^2 \times w)$} & $O(w+d)$ \\\hline  
 NM-DDM & $O(w \times d \times b)$ & $O(d \times b)$ \\\hline  
 HDDDM & $O(\lfloor \sqrt{w} \rfloor \times d)$ & $O(\lfloor \sqrt{w} \rfloor \times d)$\\\hline
 IKS & $O(log(w))$ & $O(w \times d)$  \\\hline
 BNDM & $O(log(w))$ & $O(w \times d)$  \\\hline  
 ECHO & $O(w^2)$ & $O(w)$  \\\hline  
 OMV-PHT & $O(w+b)$ & $O(w+b)$ \\\hline  
 DbDDA & \makecell[lt]{$O(1)$ \\ ens.: $O(e)$} & \makecell[lt]{$O(w)$ \\ ens.: $O(e \times w)$ }  \\\hline  
 CM & $O(w)$ & $O(w \times d)$ \\\hline  

\end{tabular}
	\label{table:complexities}
\end{table}

\autoref{table:complexities} shows the time and space complexities of the considered approaches when processing a stream.
~Only ECHO has a quadratic time complexity with respect to the window size, i.e.~doubling $w$ leads to quadrupling the runtime.~CD has a quadratic time complexity with respect to the dimensions of the data only for the case of drift as reported in the original publication. Therefore, CD's time complexity is dependent on the nature of the data but also on the quality (accuracy) of the drift detection itself, i.e. reporting too much drift, leads to a high increase of the runtime. 
KL is the only approach that depends only on the data dimensions $d$ and an internal parameter $\delta$ for maintaining the data structure incrementally.
~The other approaches' time complexities depend on the window size $w$ directly, i.e. examined data points. 
~The lowest time complexity is determined for DbDDA by iteratively maintaining the mean and standard deviation over the \gls{ml} model confidences in two windows and comparing this to a threshold.~Therefore, the time complexity is constant, and for the ensemble case, the number of ensemble members $e$.
~For the space complexity, KL, PR, IKS, BNDM and CM store all the data points, i.e. $O(w \times d)$. 
ECHO and DbDDA rely on $w$ only since only one value per window needs to be stored.
For DbDDA ensemble, space complexity grows proportionally with the ensemble size $e$ and $w$.

Complexities are important pillars in analyzing a \glspl{dd}' behavior concerning runtime and memory since they highlight factors that influence time and space complexity behaviors. However, they are not enough to assess real-world scenarios on real data since the complexities only reflect the scaling behavior but no absolute runtimes or memory requirements. Thus, even if the complexity of one approach is higher than another one, better implementation and higher drift detection accuracy might overcome this theoretical drawback. Indeed, this should point to the necessity of proper and optimized implementations, empirically verifying the estimated theoretical complexities. Additionally, the approaches' detection accuracy is not reflected in the time and space complexities at all.
Thus, we need empirical evidence reflecting on the computational performance and quality as a crucial additional investigation for concept drift detection. 

\subsection{Benchmarking}
Conducting a benchmark in the field of \gls{ml} often refers to comparing quality measures, e.g. accuracy or recall of different approaches.~Therefore, aiming for computational performance as one main objective in benchmarking faces particular challenges as initially discussed in~\cite{MattsonEtAl2020_MLPerf}.~These particular challenges are explained in this section with respect to the field of unsupervised concept drift detection.

\paragraph{Diversity in Datasets:}

The works identified in~\autoref{sec:unsupervised} leverage different synthetic and real-world datasets that are used to benchmark unsupervised \glspl{dd}. ~Synthetic data allows defining drift patterns, e.g. drift occurrence or frequency.~Moreover, it is possible to define the kind of drift, i.e. virtual vs real drift, or the temporal occurrence, e.g. abrupt, incremental, gradual, or reoccurring~\cite{webb2016characterizing}.
For a benchmark, this has the advantage of having a ground truth, we can compare drift detection results and create individual scenarios to reflect different settings for drift detection. Examples of synthetic data sources are Hyperplane~\cite{hulten2001mining} or Agrawal~\cite{agrawal1993database} among others, that are implemented in the Massive Online Analysis~\cite{bifet2010moa} framework.
~However, the effectiveness of using synthetic data depends on how well the data represents real-world characteristics. This is emphasized by \cite{souza2020challenges} explaining the importance and challenges for benchmarking against real-world data and providing a broad overview of available real-world datasets such as Abrupt Insects~\cite{dos2016fast}, Gas~\cite{VERGARA2012320}, Electricity~\cite{harries1999splice}, Forest Covertype~\cite{blackard1999comparative} or Airlines~\cite{ikonomovska2011learning}. 
This is further affirmed by~\cite{gemaque2020overview}, referring to the fact that the amount of data used for evaluating \glspl{dd} is small across the whole literature. 
~Furthermore, we recognized a big gap between the amount of data used for evaluating \glspl{dd} compared to benchmarks in the data streaming domain that use datasets with multiple million data points, e.g. Amazon movie reviews~\cite{mcauley2013amateurs} or Google web graph~\cite{leskovec2009community}. 
Additionally to data sizes, we propose to reflect on class imbalances and multi-class data. The multi-class scenario is given by many of the real-world datasets, but should also be considered in synthetic scenarios explicitly. \cite{palli2022experimental} give evidence on different behaviors of supervised \glspl{dd} for class imbalances in comparison to evenly distributed class cases. They report similar results for multi-class data in comparison to binary class. We need to extend this research for unsupervised \glspl{dd}.

\paragraph{Model Dependency:}

According to the survey by~\cite{shen2023unsupervised}, \glspl{dd} can be separated into two groups: A) differences in data distribution and B) model quality monitoring.~The latter can be considered model-dependent since it directly relies on a model's quality, whereas the first group can be considered model-independent.
For the model-dependent approaches, we identify two cases of unreliable drift detection.
The first case is when the model reports high inference quality (e.g.~high confidence) after concept drift, leading to wrong predictions but no detection by the \gls{dd}. Even if there are no studies that prove this behavior in general, there are studies that prove it for selected models.~E.g.~\cite{nguyen2015deep} provides evidence that deep learning models produce high confidence for incorrect predictions, e.g.~after concept drift. \cite{lughofer2016recognizing} shows, that this is also the case for evolving fuzzy classifiers. 
The second case is when the model reports low inference quality (e.g. low confidences) without concept drift, leading to incorrectly detected drift and unnecessary computational overhead due to subsequent drift handling strategies. Overfitting~\cite{hawkins2004problem} could be a reason for such behavior, which applies to all types of \gls{ml} models.
Thus, we conclude, due to the heterogeneity of \gls{ml} models, it is important to reflect on different base models for a benchmark comprehensively comparing \glspl{dd} from groups A~and~B.

\begin{figure*}[t]
 \centering
\includegraphics[width=\textwidth]{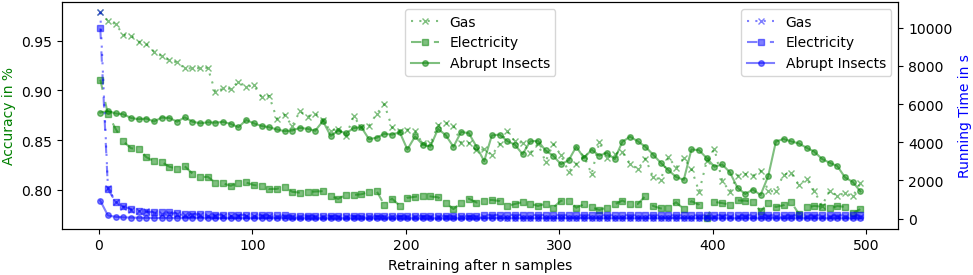}
\includegraphics[width=\textwidth]{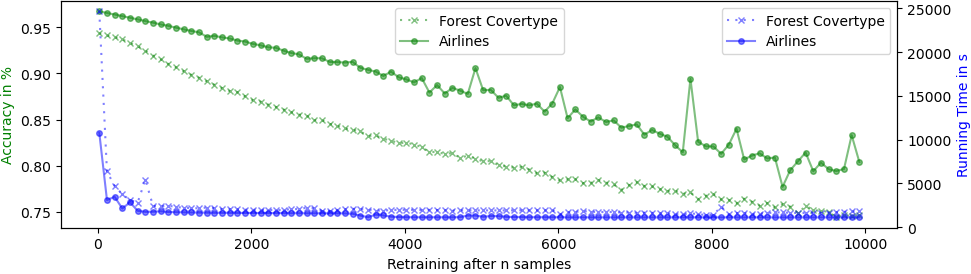}
    \caption{Accuracy and runtime behavior while processing streams over different datasets. The X-axis indicates the number of samples after which re-training is conducted. The Y-axis (right/blue) shows runtime and Y-axis (left/green) shows accuracy. \textit{Gas, Electricity, Abrupt Insects} where initialized with $n=1$ and sampled in steps of 5, \textit{Forest Covertype, Airlines} were initialized with $n=20$ and sampled in steps of 100.}
    \label{fig:baselines}
\end{figure*}

\paragraph{Baselines:}

The literature often suggests \textit{no concept drift detection} as a baseline for a benchmark~\cite{souza2021efficient,cerqueira2022studd,dos2016fast}. In particular, this refers to an initially trained \gls{ml} model that is not retrained during the processing of a data stream, nor is any action beyond the inference process considered.~Therefore, this is the best possible runtime for a given \gls{ml} model and data stream and does not require any further labeled data. On the other hand, this baseline indicates the minimum inference quality (e.g. accuracy) that a pipeline with a \gls{dd} should achieve.
Thus, a second baseline should reflect the best inference quality possible while processing the stream by re-training the \gls{ml} model after $n$ samples. ~Therefore, it requires $100\%$ of the true labels. \cite{souza2021efficient} defines $n=1$ and retrains the \gls{ml} model after each sample. This results in a very high runtime overhead for that pipeline due to the permanent re-training of the \gls{ml} model but guarantees the highest possible accuracy. 

Additionally, we propose to consider at least one more baseline.
This is due to the asymptotic behavior of the runtime and the differing behavior of the inference quality, e.g. accuracy, while increasing $n$ that indicates the number of samples after which a classifier is re-trained while processing a stream. To show this, we processed the five introduced datasets: Abrupt Insects, Gas, Electricity, Forest Covertype and Airlines. We implemented a Random Forest classifier, that is trained on initial data and updated every $n$ samples. For the smaller datasets Abrupt Insects, Gas, and Electricity we used 500 samples for the initial training, while for the bigger Forest Covertype and Airlines datasets, we used 2000 samples.~We initialized $n=1$ and increased $n$ by 5 for each run while processing the small datasets. For the big ones, we initialized $n=20$ and increased $n$ by 100.
Our results are depicted in \autoref{fig:baselines}. For all datasets, the runtime strongly decreases while increasing $n$.~Accuracy decreases slowly for all datasets and higher $n$ might have a higher accuracy than lower $n$ depending on the dataset and how well the period of re-training matches the drift pattern. In fact, we suggest a value of $n$ such that the runtime is low, while still maintaining the high accuracy provided at the bend of the runtime curve for all datasets.

\cite{cerqueira2022studd} defines $n=100$, still achieving high inference quality on their considered datasets while heavily reducing the runtime overhead since re-training is conducted only for each $100$ samples. While this is a good starting point for a pragmatic baseline, it might not be generally applicable since it does not necessarily reflect on an optimum of the curves, balancing accuracy and runtime.
~However, one can explore alternative values for $n$, i.e. baselines depending on the dataset's characteristics.

\paragraph{Set of Metrics:}
In~\cite{barros2018large}, metrics to track the inference quality of a supervised \gls{dd} such as Accuracy, Precision, Recall, F1-Score, True Positives, False Positives, Detection Delay, and Mathew's Correlation Coefficient were used. Additionally, we propose to track the (cumulative) Accuracy Gain per Drift as presented by~\cite{wu2021nacre} to measure the effectiveness of a \gls{dd} throughout a stream and not only once for the whole dataset. For unsupervised \gls{dd} we also suggest tracking the amount of requested labels, since some \glspl{dd} detect drift in an unsupervised manner but request labels for subsequent actions. 
To track computational performance, we propose metrics such as runtime, resource utilization, and efficiency, i.e.~memory or CPU utilization. ~\cite{mahgoub2022benchmarking} did a similar investigation and tracked the runtime and memory of several supervised \glspl{dd}. As explained in~\cite{MattsonEtAl2020_MLPerf}, inference quality and computational performance can't be considered independently, since achieving higher quality might also require further computational overhead as we also showed in \autoref{fig:baselines}. Thus, we propose two strategies: 1.) select a threshold or set a target quality for the inference, and measure the computational resources to achieve that 2.) set a threshold for the runtime or memory and compute the inference quality (accuracy) that is achievable under this resource constraints. As indicated by \autoref{table:complexities}, a similar runtime for two \glspl{dd} might require different settings for \gls{dd} parameters, e.g. window size $w$.

We highlighted four main considerations required for a comprehensive benchmark of unsupervised \glspl{dd}.~However, one preliminary requirement is the availability of proper implementations of the \glspl{dd}, i.e. without crucial performance bugs that deteriorate the computational performance of the implementations as outlined by~\cite{jin2012understanding}. Nevertheless, as investigated by~\cite{lukats2023reproducible}, implementations are scarce, and prose details are often insufficient for proper re-implementations. This should point to the necessity of the provision of proper descriptions or implementations of the \glspl{dd} by the community.

\begin{figure*}
    \centering
\includegraphics[width=\textwidth]{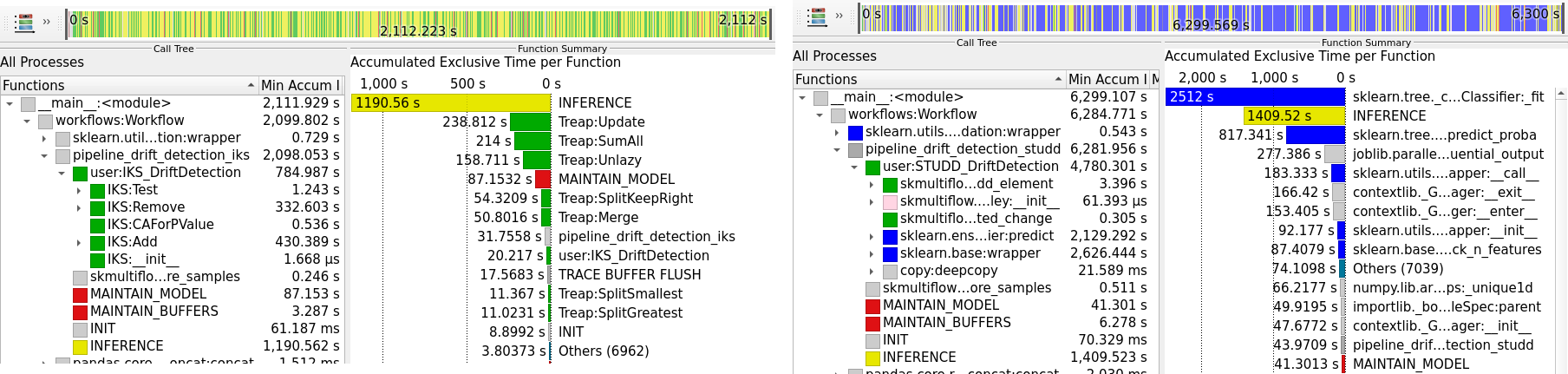}
    \caption{Vampir display of the IKS \gls{dd} (left) and STUDD \gls{dd} (right) performance data on the Forest Covertype dataset. 
    In each display, the left part shows the call tree and the right part shows the function summary, i.e. accumulated exclusive time per function.
    User regions for the inference process are yellow, user regions for the drift detection are green, user regions for maintaining buffers, and the \gls{ml} model is red. \texttt{sklearn} functions are blue. The remaining functions are shaded grey.}
    \label{fig:traceStudd}
\end{figure*}

\subsection{Performance Analysis}

Performance analysis refers to testing, analyzing, and optimizing systems to achieve computational performance goals and to identify and resolve performance bugs~\cite{jin2012understanding}. It usually follows a two-step workflow: 1.) profiling, i.e. accumulating performance data, e.g. runtime of application functions or memory 2.) tracing of an application to exactly record program regions enter and exit over time. Both steps support a performance analyst in identifying computational bottlenecks in implementations and improving an application's behavior. For data science, it is also crucial to reflect on the quality (e.g. accuracy) when conducting performance analysis, emphasizing the importance of holistic performance engineering, e.g. improving the runtime behavior of an implementation while maintaining benchmark results.
Score-P~\cite{knupfer2012score} is an established, scalable open-source framework that supports profiling and tracing.
~With the Score-P Python bindings~\cite{gocht2021advanced}, it supports the Python programming language, which can be considered predominant in data science. ~Additionally, a Score-P Jupyter kernel~\cite{werner2021bridging} makes it available for execution in Jupyter notebooks.
Vampir~\cite{vampir} is a tool to display performance data as collected by Score-P.

To demonstrate the effectiveness of these workflows, we measured the run of the two Python-based \glspl{dd} STUDD and IKS on the Forest Covertype dataset. ~\autoref{fig:traceStudd} displays the performance data of the STUDD and IKS run in Vampir.
~The left part shows the call tree displaying the invocation hierarchy across the functions and accumulates the runtime. The middle part displays a timeline of the called functions (top) and their call stack (bottom). The right part shows the function summary, i.e. accumulated exclusive time per function.
For both displays, we consider user regions for the inference process (yellow), user regions for the drift detection (green), user regions for maintaining buffers, and the \gls{ml} model (red). The other functions are shaded grey. 

For STUDD, we additionally colored \texttt{sklearn} functions in blue and derived the following:   
As displayed in the function summary, the STUDD implementation mostly relies on \texttt{sklearn} while processing the stream. Based on the call tree, we see that the time for drift detection ($4780s$, $76\%$) predominates over time for inference ($1409s$, $22\%$). The call tree also shows the high dependency of \texttt{user:STUDD\_DriftDetetion} on the \texttt{sklearn} base functionalities.~With these insights, we conclude that the runtime overhead for STUDD is significantly high for processing the stream only due to the drift detection.~This is mainly due to the approach of applying and maintaining an additional student model. However, since it is mostly based on \texttt{sklearn}, which is an established library with additional support for parallel execution, it is worth investigating how STUDD scales across multiple cores or nodes.

\begin{figure*}
    \centering
\includegraphics[width=\textwidth]{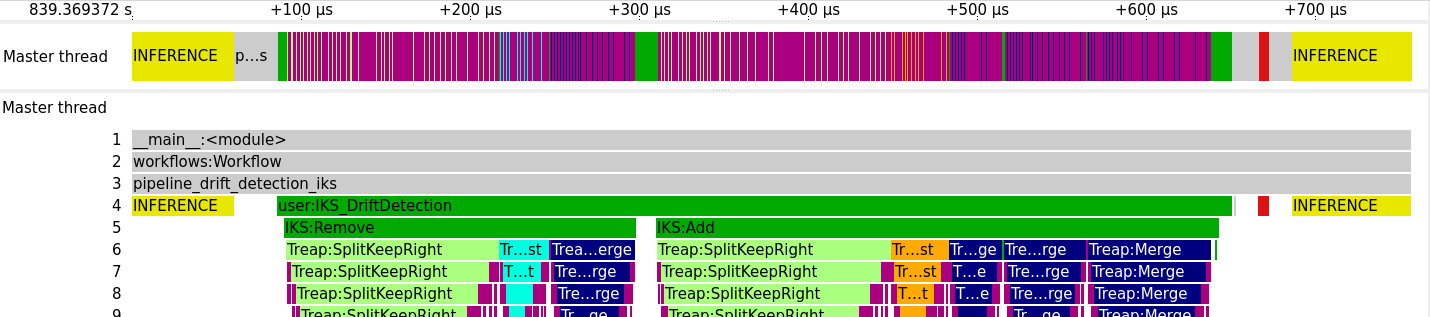}
    \caption{Vampir display of the IKS \gls{dd} with timeline feature, i.e. chronological sequence of the called functions. We selected the time window to show the called functions of the IKS per sample, i.e. between inference steps (yellow). The top display shows the master timeline and the bottom display the call stack.
    \texttt{Treap.SplitKeepRight} is light green, \texttt{Treap.KeepGreatest} cyan, \texttt{Treap.KeepSmallest} orange, \texttt{Treap.Merge} dark blue and other \texttt{Treap} functions are purple.}
    \label{fig:traceIks}
\end{figure*}

For IKS, the most time for processing the pipeline was spent on the inference process ($1191s$, $56\%$) as shown in the call tree and the function summary. Nevertheless, drift detection also takes $784s$ ($37\%$) as the call tree node \texttt{user:IKS\_DriftDetetion} shows. This overhead is mostly based on the \texttt{Add} and \texttt{Remove} functions of the IKS implementation, e.g.~maintaining the internal \textit{Treap} data representation.~The time for maintaining the \gls{ml} model and buffers can be considered insignificant. ~With Vampir's timeline chart, it is possible to get further insights into chronological sequences of called functions. In \autoref{fig:traceIks} we selected the period between the inference of two subsequent samples of the data stream to investigate IKS' behavior per sample.~We can see the recursive call within \texttt{IKS:Remove} for removing the oldest sample in the reference window.~Subsequently, the current sample is added and triggers a recursive call within \texttt{IKS:Add}.~Adding and removing a sample mainly consists of a similar recursive sequence of methods, i.e. \texttt{Treap:SplitKeepRight} (light green) at the beginning, \texttt{Treap:Merge} (dark blue) at the end and either \texttt{Treap:KeepGreatest} (cyan) or \texttt{Treap:KeepSmallest} (orange) in between. \texttt{IKS:Add} consists of additional merges. 

Therefore, we conclude that IKS mostly relies on the internal data representation. The resource-efficient and potentially parallel implementation of maintaining this representation is the key to a low runtime overhead.
While we did not perform an in-depth performance analysis of IKS and STUDD, we still show the effectiveness of such investigations for \glspl{dd}. Further analysis should follow to support the development of resource-efficient scalable \glspl{dd}.

\section{\uppercase{Conclusion}}
This work contributes to computational performance engineering for unsupervised \glspl{dd} by discussing computational complexities, a comprehensive benchmark, and an initial performance analysis of two \glspl{dd}.
We show the necessity of such investigation by highlighting the gap in the prior evaluation and demonstrating a high runtime of two \glspl{dd} on a larger dataset.
Although time and space complexities alone are not enough for comprehensive performance engineering, they are an important pillar for such investigations. Thus, we determined them for existing approaches. The complexities can be reflected in a benchmark that takes into account the important aspects we have discussed, i.e.~diversity of datasets, model dependencies, baselines, and the set of metrics. 
~Performance analysis provides in-depth insights into application behaviors and supports the development of resource-efficient and parallel \glspl{dd}. We provide an initial analysis with the tools Score-P and Vampir to reveal potential bottlenecks in the implementations of two \gls{dd}. 

For future work, we plan to extend our complexity analysis and substantiate the determined time and space complexities with empirical measurements. Moreover, we want to provide a comprehensive benchmark of unsupervised \glspl{dd} reflecting the computational performance and inference quality.~Using gained insights, we aim to develop parallel and resource-efficient solutions for diverse applications.

\section*{\uppercase{Acknowledgment}}
The authors gratefully acknowledge the computing time made available to them on the high-performance computer at the NHR Center of TU Dresden. This center is jointly supported by the Federal Ministry of Education and Research and the state governments participating in the NHR (www.nhr-verein.de/unsere-partner).

\bibliographystyle{apalike}
\bibliography{main.bib}

\section*{\uppercase{Appendix}}

We justify the determined time and space complexities. All approaches refer to at least two parameters: $w$ - window size (number of examined data points) and $d$ - number of data dimensions per data point\\

\textbf{KL~\cite{dasu2006information}}
$\delta$ - minimum value for the property \textit{box length} of their internal data representation \textit{kdq-tree}
\quad State the time and space complexity in the publication.

\textbf{PR~\cite{gu2016concept}}
State the time complexity of their approach in the publication.
For the space complexity, they store $w$ instances with $d$ features, i.e. $O(w*d)$

\textbf{CD~\cite{qahtan2015pca}}
Even not mentioned explicitly in their publication, processing a sample relies on the data dimensionality $d$ since the projection of a data point onto the PCs is done in $O(d \times k)$, where $k$ is the number of PCs describing $99.9\%$ of the variance and $k$ can be considered as a constant. For drift, they report $O(d^2 \times w)$.
Space complexity for the PCs with a fixed $k$ is $O(d)$ and $O(w)$ due to the projection of the data points in the reference window on a fixed number of PCs, i.e. $O(w+d)$.

\textbf{NM-DDM~\cite{mustafa2017unsupervised}}
$b$ - number of bins of estimated histogram
\quad State the time complexity of their approach in the publication.
For the space complexity, $b$ number of bin values in each of the histograms for the $d$ features are stored, i.e. $d*b$.

\textbf{HDDDM~\cite{ditzler2011hellinger}}
Originally HDDDM was proposed as a batch-based approach. To apply it incrementally on each sample, we assume a naive procedure: we introduce an initialization phase over the first samples $S_i$ with $i=1,2,...w$ and initialize $D_\lambda = D_w = {S_1,...S_w}$. For each new sample $S_t$ and $t=w+1, w+2,...$, we initialize $D_t=D_{t-1} \cup {S_t} \\ {S_{t-w}}$ to create a sliding window of size $w$. Now, we apply the HDDDM as usual.

Drift detection is performed by Hellinger distances between two histograms based on $S_\lambda$ and $S_t$. The histograms have $\lfloor \sqrt{w} \rfloor$ bins and the main computational complexity is in the Hellinger distance between these two histograms, i.e. $O(\lfloor \sqrt{w} \rfloor \times d)$ to compute the intermediate distances per bin of each feature, resulting in the final Hellinger distance.
For space complexity, only the histograms must be kept in memory, i.e. $O(\lfloor \sqrt{w} \rfloor \times d)$.

\textbf{IKS~\cite{dos2016fast}}
State the time complexity of their approach in the publication.
For the space complexity, they store $w$ instances with $d$ features, i.e. $O(w*d)$

\textbf{BNDM~\cite{xuan2020bayesian}}
State the time complexity of their approach in the publication.
For the space complexity, they store $w$ instances with $d$ features, i.e. $O(w*d)$

\textbf{ECHO~\cite{haque2016efficient}}
State the time and space complexity of their approach in the publication.

\textbf{OMV-PHT~\cite{lughofer2016recognizing}}
$b$ - number of bins of estimated histogram
\quad
Propose an extension of the Page-Hinkley test by only considering deteriorations of the quality indicator (e.g. accuracy) and fading older samples for computing the quality indicator. Besides a supervised quality indicator, they propose a semi-supervised one, that relies on active user feedback, and an unsupervised quality indicator which is considered in this study.
To determine the quality of a model, they build histograms over the model confidence values during inference. Next, they compare the histograms of two windows by iterating over the bins, i.e. $O(b)$, and apply their faded Page-Hinkley test in $O(w)$ since the fading of older samples needs to be re-computed. An update of the histograms can be done per sample in $O(1)$. Therefore, the overall time complexity is $O(w+b)$. Space complexity is $O(w+b)$ to store the confidence histogram and the $w$ magnitude values for the OMV-PHT test.

\textbf{DbDDA~\cite{kim2017efficient}}
Calculates mean and standard deviation for a reference and detection window of size $w$ over the confidences of the \gls{ml} model in the inference. Mean and standard deviation are computed iteratively in $O(1)$ per dimension with space complexity $O(w)$ to store all confidences.
The reference window is either fixed or moving for DbDDA.
For the ensemble case, multiple reference windows are kept, and mean and standard deviation are considered w.r.t. this ensemble, i.e. time complexity $O(e)$ and space complexity $O(e*w)$.

\textbf{CM~\cite{lu2014concept}}
State the time complexity of their approach in the publication.
For the space complexity, they store $w$ instances with $d$ features, i.e. $O(w*d)$

\end{document}